\documentclass[conference]{IEEEtran}
\usepackage[utf8]{inputenc}
\usepackage{verbatim}
\usepackage[cmex10]{amsmath}
\usepackage{subfigure}
\usepackage{booktabs}
\usepackage{graphicx}
\usepackage{pifont} 
\newcommand{\tick}{\ding{52}}

\IEEEoverridecommandlockouts

\date{}
\title{Extending Similarity Measures of Interval Type-2 Fuzzy Sets to General Type-2 Fuzzy Sets\thanks{This work was partially funded by the EPSRC’s Towards Data-Driven Environmental Policy Design grant, EP/K012479/1 and the RCUK’s Horizon Digital Economy Research Hub grant, EP/G065802/1.}}
\author{
  \IEEEauthorblockN{Josie McCulloch}
  \IEEEauthorblockA{
    Student Member\\
    School of Computer Science\\
    University of Nottingham\\
    Nottingham, United Kingdom\\
    psxjm5@nottingham.ac.uk}
\and
  \IEEEauthorblockN{Christian Wagner}
  \IEEEauthorblockA{
    Member \\
    School of Computer Science\\
    University of Nottingham\\
    Nottingham, United Kingdom\\
    christian.wagner@nottingham.ac.uk}
\and
  \IEEEauthorblockN{Uwe Aickelin}
  \IEEEauthorblockA{School of Computer Science\\
  University of Nottingham\\
  Nottingham, United Kingdom\\
  uwe.aickelin@nottingham.ac.uk}
}

\begin{document}
\maketitle
\begin{abstract}
Similarity measures provide one of the core tools that enable reasoning about fuzzy sets. While many types of similarity measures exist for type-1 and interval type-2 fuzzy sets, there are very few similarity measures that enable the comparison of general type-2 fuzzy sets. In this paper, we introduce a general method for extending existing interval type-2 similarity measures to similarity measures for general type-2 fuzzy sets. Specifically, we show how similarity measures for interval type-2 fuzzy sets can be employed in conjunction with the zSlices based general type-2 representation for fuzzy sets to provide measures of similarity which preserve all the common properties (i.e. reflexivity, symmetry, transitivity and overlapping) of the original interval type-2 similarity measure. We demonstrate examples of such extended fuzzy measures and provide comparisons between (different types of) interval and general type-2 fuzzy measures.
\end{abstract}


\section{Introduction}
Fuzzy logic has been successfully applied to many real world applications in which uncertainty is present. Type-1 (T1) fuzzy logic has been the most popular form of fuzzy logic used, however, advances in fuzzy logic theory have made it possible to research more complex types of fuzzy logic such as interval type-2 (T2) and general T2 fuzzy sets and systems. In particular, interval T2 fuzzy sets (FSs) have been used intensively because their computational requirements compared to that of general T2 FSs are greatly reduced.

One of the most common tools of fuzzy logic is similarity measures (SMs). A SM between FSs indicates the degree to which the FSs are similar. The concept is relevant in many fields, for example, pattern recognition \cite{Mitchell2005}, analogical reasoning \cite{similarityreasoning} and fuzzy rule base simplification \cite{setnes1998similarity}. SMs for T1 FSs have been extensively studied by many researchers, such as  \cite{Liu200561}, \cite{Lee_designof} and \cite{Zwick1987} where the latter provides a good overview. However, SMs for T2 FSs have been less widespread. Although some methods have been developed for interval T2 FSs, e.g. \cite{Zeng20061477, WuComparative, Gorzalczany1987, Bustince2000137, WuVector}, fewer methods exist for general T2 FSs.

Four properties of SMs for FSs that are commonly used in the literature are:\\
Reflexivity: $s(\tilde{A}, \tilde{B}) = 1 \Longleftrightarrow \tilde{A} = \tilde{B} $\\
Symmetry: $s(\tilde{A}, \tilde{B}) = s(\tilde{B}, \tilde{A}) $\\
Transitivity: If $\tilde{A} \leq \tilde{B} \leq \tilde{C}$, then $s(\tilde{A},\tilde{B}) \geq s(\tilde{A}, \tilde{C})$\\
Overlapping: If $\tilde{A} \cap \tilde{B} \neq \emptyset$, then $s(\tilde{A}, \tilde{B}) > 0$; otherwise, $s(\tilde{A}, \tilde{B}) = 0$\\
Note that it is not necessary for a SM to have all of these properties as the application of the measure may not depend on all of them.

With the recent increase in T2 applications there is a growing potential for applications of T2 SMs.
This paper presents a general method of extending SMs on interval T2 FSs to SMs on general T2 FSs. 
In Section II, background theory of T1 and T2 FSs will be presented, followed by an overview of some of the most common existing methods of similarity for interval T2 FSs. In Section III the generalisation of interval T2 SMs to general T2 FSs will be introduced, followed by demonstrations and comparisons of the newly introduced SM for general T2 FSs and existing interval T2 SMs in Section IV. Finally, some conclusions are presented in Section V.

\section{Background}
\subsection{Fuzzy Sets}
\subsubsection{Type-1 Fuzzy Sets}
T1 FSs have been applied to many fields from data mining \cite{datamining} to time-series prediction \cite{timeSeries} and computing with words \cite{CWW}.
The T1 FS $F$ is represented as
\begin{equation}
 F = \int_X \mu_F(x)/x
  \label{type1eq}
\end{equation}
where $\int$ denotes the collection of all points $x \in X$ with associated membership function (MF) $\mu_F(x)$ \cite{mendel2001uncertain}.

\subsubsection{General Type-2 Fuzzy Sets}
\begin{figure}
  \centering
    \subfigure[]
    {
      \includegraphics[scale=0.2]{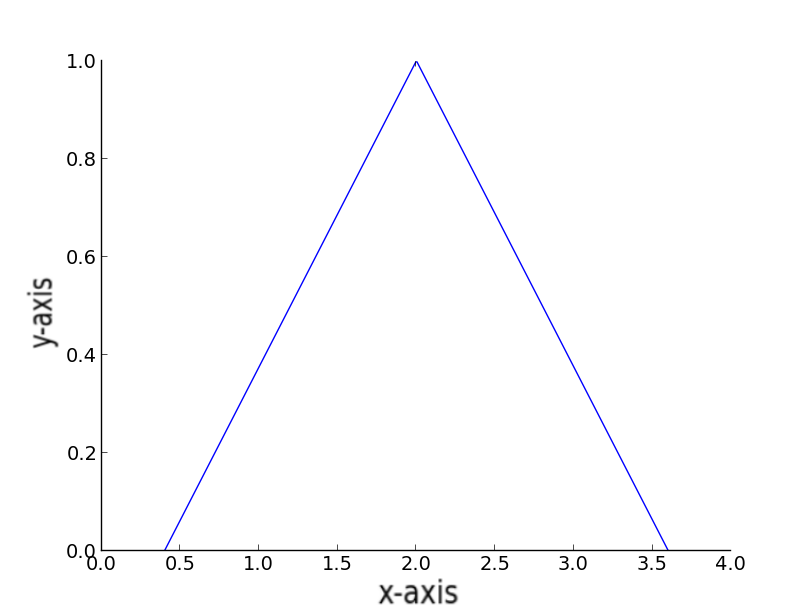}
    }
    \subfigure[]
    {
      \includegraphics[scale=0.2]{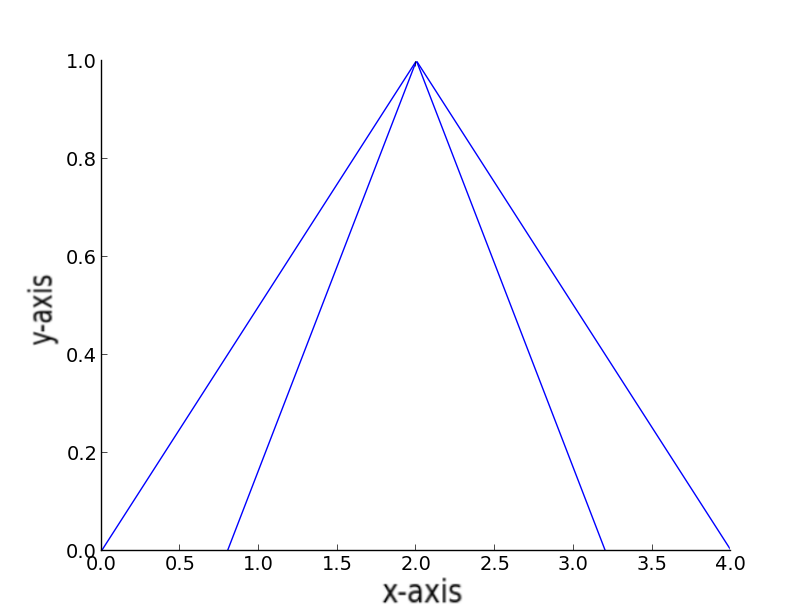}
    }
  \caption{(a) A type-1 fuzzy set. (b) A type-2 fuzzy set.}
  \label{fig:simpleFSs}
\end{figure}

A T2 FS differs from a T1 FS in that it has a Footprint Of Uncertainty (FOU) which is defined by two MFs; a lower MF and an upper MF. An example of a T1 and T2 FS can be seen in Fig. \ref{fig:simpleFSs}. For any input, $x$, in the T1 FS the membership value is a crisp number in [0,1], whereas for the T2 FS the membership value is a T1 FS. A general T2 FS $\tilde{F}$ can be expressed as
\begin{equation}
\begin{array}{l r}
 \tilde{F} = \int_{x \in X} \int_{u \in J_x} \mu_{\tilde{F}}(x,u) / (x, u) & J_x \subseteq [0,1]
\end{array}
\end{equation}
where $x$ is the primary variable in $X$, $u$ is the secondary variable which has the domain $J_x \subseteq [0,1]$, and the amplitude of $\mu_{\tilde{F}}(x,u)$ is known as the secondary grade. 

\subsubsection{Interval Type-2 Fuzzy Sets}
An interval T2 FS is a special case of a general T2 FS in which the secondary grade equals 1 for $\forall x \in X$ and $\forall u \in J_x $. Thus the interval T2 FS $\tilde{F}$ can be expressed as \cite{ITFSsimple}
\begin{equation}
\begin{array}{l r}
 \tilde{F} = \int_{x \in X} \int_{u \in J_x} 1 / (x, u) & J_x \subseteq [0,1]
\end{array}
\end{equation}
Interval T2 FSs are the most commonly used T2 FSs because of their simplicity and reduced computational cost in comparison to general T2 FSs \cite{WuFundamentalDifferences}.

T2 FSs have been successfully applied to many fields, such as autonomous mobile robots \cite{T2AutonomousRobot}, decision making \cite{typeIIdecisions} and forecasting of time-series \cite{Karnikforecasting}. Additionally, in recent years T2 FSs have proven to outperform T1 FSs in many applications (when the number of sets is kept constant) because they are able to model uncertainty with greater accuracy \cite{T2AutonomousRobot, T2simple, experimentalstudy}.

\subsubsection{zSlices-Based General Type-2 Fuzzy Sets}
Many efforts have been made to reduce the complexity of general T2 FSs. Coupland and John introduced a geometric representation of T2 FSs \cite{CouplandJohnGeometric}, Mendel et al. proposed the use of alpha-planes \cite{5067334} and Wagner and Hagras put forward the zSlices approach \cite{zSlicesPaper2}. This paper will focus on the use of zSlices, which are described next.

A general T2 FS can be represented by slicing the third dimension ($z$) at level $z_i$ to create a zSlices-based general T2 FS \cite{zSlicesPaper2}. The resulting set will consist of zSlices which are interval T2 FSs with a secondary membership grade of $z_i$; unlike regular interval T2 FSs, whose secondary membership grade is always 1. Thus, the zSlice $\tilde{Z}_i$ can be written as follows \cite{zSlicesPaper2}:
\begin{equation}
  \tilde{Z}_i = \int_{x \in X} \int_{u_i \in J_{i_x}} z_i / (x, u_i).
\end{equation}
The FS $\tilde{F}$ can then be represented as a collection of zSlices \cite{zSlicesPaper2}:
\begin{equation}
  \tilde{F} = \sum_{i=1}^I \tilde{Z}_i
\end{equation}
where $I$ is the number of zSlices. Note that the zSlice $Z_0$ is disregarded because its secondary grade is 0, and thus $Z_0$ does not contribute to the FS \cite{zSlicesPaper2}. 
As the number of zSlices used to represent a general T2 FS increases, the zSlices-based T2 FS's representation of the original set becomes more accurate.
In addition to the purpose of simplifying general T2 FSs, ``pure'' zSlices-based general T2 FSs have also been used in the literature to model agreement \cite{zsliceAgreement}, \cite{zslicesAgreementTwo}, for example in the context of Computing With Words \cite{CWW, mendel2010perceptual}.

\subsection{Similarity Measures}
In this section, a collection of SMs for interval T2 FSs will be briefly reviewed. For conciseness we focus on four existing methods for measuring the similarity between two interval T2 FSs. Though additional measures exist, the measures presented in this section are the most common and provide a good overview of the interval T2 SMs available. A more detailed comparison of these and other interval T2 SMs has been presented by Wu and Mendel in \cite{WuVector} and \cite{WuComparative}.

\subsubsection{Zeng and Li}
Zeng and Li \cite{Zeng20061477} proposed the following SM for interval T2 FSs, $\tilde{A}$ and $\tilde{B}$, in a discrete universe of discourse:
\begin{equation}
  \begin{array}{l}
  S_{ZL}(\tilde{A},\tilde{B}) = \\
    1 - \frac{1}{2n} \sum^n_{i=1} 
      ( |\underline{\mu}_{\tilde{A}}(x_i) - \underline{\mu}_{\tilde{B}}(x_i)| + 
      |\overline{\mu}_{\tilde{A}}(x_i) - \overline{\mu}_{\tilde{B}}(x_i)| )
  \end{array}
  \label{zeng}
\end{equation}
where $ \underline{\mu}_{\tilde{A}}(x_i) $ denotes the lower membership value of the FS $\tilde{A}$ at $x_i$, $ \overline{\mu}_{\tilde{A}}(x_i) $ denotes the upper membership value of $\tilde{A}$ at $x_i$, and n is the total number of discretisations along the $x$-axis.
This method has the properties of reflexivity, symmetry and transitivity. However, it does not have the property of overlapping, and instead as the distance between two disjoint sets increases their similarity according to the measure also increases.

\subsubsection{Jaccard}
Wu and Mendel \cite{WuComparative} and Nguyen and Kreinovich \cite{kreinovich} proposed the following SM for interval T2 FSs $\tilde{A}$ and $\tilde{B}$:
\begin{equation}
  \begin{array}{l}
    S_J(\tilde{A},\tilde{B}) = \\
      \frac{
	\int_X min(\overline{\mu}_{\tilde{A}}(x), \overline{\mu}_{\tilde{B}}(x)) dx + 
	\int_X min(\underline{\mu}_{\tilde{A}}(x), \underline{\mu}_{\tilde{B}}(x)) dx}
	{
	\int_X max(\overline{\mu}_{\tilde{A}}(x), \overline{\mu}_{\tilde{B}}(x)) dx + 
	\int_X max(\underline{\mu}_{\tilde{A}}(x), \underline{\mu}_{\tilde{B}}(x)) dx}
    \end{array}
	\label{jaccard}
\end{equation}
This method has the properties of reflexivity, symmetry, transitivity and overlapping.

\subsubsection{Gorza{\l}czany}
Gorza{\l}czany proposed a compatibility measure for two interval T2 FSs \cite{Gorzalczany1987}; a compatibility measure has been described as a broad concept which typically encompasses both similarity and proximity \cite{cross2002similarity}. Gorza{\l}czany's measure is given as follows:
\begin{subequations}
  \begin{equation}
    S_G(\tilde{A},\tilde{B}) =  
       [S^L(\tilde{A},\tilde{B}),\ S^U(\tilde{A},\tilde{B})]
  \end{equation}
  \begin{equation}
    S^L(\tilde{A},\tilde{B}) =  min\{S_1(\tilde{A},\tilde{B}), S_2(\tilde{A},\tilde{B})\}
  \end{equation}
  \begin{equation}
    S^U(\tilde{A},\tilde{B}) =  max\{S_1(\tilde{A},\tilde{B}), S_2(\tilde{A},\tilde{B})\}
  \end{equation}
  \begin{equation}
    S_1(\tilde{A},\tilde{B}) = 
       \frac{
          max_{x \in X}(min[\underline{\mu}_{\tilde{A}}(x), \underline{\mu}_{\tilde{B}}(x)])}
          {max_{x \in X}\underline{\mu}_{\tilde{A}}(x)}
  \end{equation}
  \begin{equation}
    S_2(\tilde{A},\tilde{B}) = 
       \frac{
          max_{x \in X}(min[\overline{\mu}_{\tilde{A}}(x), \overline{\mu}_{\tilde{B}}(x)])}
          {max_{x \in X}\overline{\mu}_{\tilde{A}}(x)}
  \end{equation}
  \label{Gorz}
\end{subequations}
Wu and Mendel have shown in \cite{WuVector} that this measure does not reflect reflexivity. The result of $S_G(\tilde{A},\tilde{B})$ is always (1, 1) when $max_{x \in X}\underline{\mu}_{\tilde{A}}(x) = max_{x \in X}\underline{\mu}_{\tilde{B}}(x)$ and $max_{x \in X}\overline{\mu}_{\tilde{A}}(x) = max_{x \in X}\overline{\mu}_{\tilde{B}}(x)$, even if the shapes of the two sets are different.

\subsubsection{Bustince}
Bustince proposed the following SM \cite{Bustince2000137}:
\begin{subequations}
  \begin{equation}
    S_B(\tilde{A},\tilde{B}) = [S_L(\tilde{A},\tilde{B}), S_U(\tilde{A},\tilde{B})]
  \end{equation}
  \begin{equation}
    S_L(\tilde{A},\tilde{B}) = 
      \Upsilon_L(\tilde{A},\tilde{B}) \star \Upsilon_L(\tilde{B},\tilde{A})
  \end{equation}
  \begin{equation}
    S_U(\tilde{A},\tilde{B}) = 
      \Upsilon_U(\tilde{A},\tilde{B}) \star \Upsilon_U(\tilde{B},\tilde{A})
  \end{equation}
  \begin{equation}
    \begin{array}{l}
      \Upsilon_L(\tilde{A},\tilde{B}) = \\
	\inf_{x \in X}\{1, min(
	  1 - \underline{\mu}_{\tilde{A}}(x) + \underline{\mu}_{\tilde{B}}(x),
	  1 - \overline{\mu}_{\tilde{A}}(x) + \overline{\mu}_{\tilde{B}}(x))\}
      \end{array}
  \end{equation}
  \begin{equation}
    \begin{array}{l}
      \Upsilon_U(\tilde{A},\tilde{B}) = \\
	\inf_{x \in X}\{1, max(
	  1 - \underline{\mu}_{\tilde{A}}(x) + \underline{\mu}_{\tilde{B}}(x),
	  1 - \overline{\mu}_{\tilde{A}}(x) + \overline{\mu}_{\tilde{B}}(x))\}
      \end{array}
  \end{equation}
  \label{Bustince}
\end{subequations}
where $\star$ is any t-norm.
This method does not support overlapping; when $\tilde{A}$ and $\tilde{B}$ are disjoint the result of  $S_B(\tilde{A},\tilde{B})$ is always greater than zero.

\section{Similarity Measures for General Type-2 Fuzzy Sets}
This section proposes a general method of extending interval T2 SMs to general T2 FSs through the zSlices representation \cite{zSlicesPaper2}. Any properties that hold for the original interval T2 SM are upheld in the general T2 SM as proven in Section B.

\subsection{Extending interval-based similarity measures}
As has been shown in \cite{zSlicesPaper2} and Section II, a general T2 FS can be represented as a series of zSlices. From this, we present the following definition of a SM for zSlices-based general T2 FSs:
\newtheorem{mydef}{\bf{Definition}}
\begin{mydef}[General type-2 similarity measure]
 By using zSlices-based general T2 FSs, a measure of similarity on interval T2 FSs can be applied to each zSlice, and the results for each zSlice can be combined as follows:
  \begin{equation}
  S_{ZS}(\tilde{A},\tilde{B}) = 
    \frac
      {\sum_{i \in L}  z_i S_\lambda(\tilde{A}_{z_i},\tilde{B}_{z_i})}
      {\sum_{i \in L}  z_i}
   \label{zSliceSM}
\end{equation}
\end{mydef}
where $S_\lambda(\tilde{A}_{z_i},\tilde{B}_{z_i})$ is any SM for interval T2 FSs. Sets $\tilde{A}_{z_i}$ and $\tilde{B}_{z_i}$ are zSlices from sets $\tilde{A}$ and $\tilde{B}$ at zLevel $z_i$, and $L$ is the set of zLevels used by $\tilde{A}$ and $\tilde{B}$. For example, if $\tilde{A}$ and $\tilde{B}$ have three zLevels where $z_1$ = 0.33, $z_2$ = 0.66 and $z_3$ = 1 then $L = \{0.33, 0.66, 1\}$.
The higher the number of zSlices that are used to represent the FSs, the more accurate the representation of the FSs, and thus the more accurate the SM will be.
When the sets use only one zLevel the equation reduces to the corresponding interval T2 SM.

\begin{figure}
  \centering
  \subfigure[]
    {
      \includegraphics[scale=0.25]{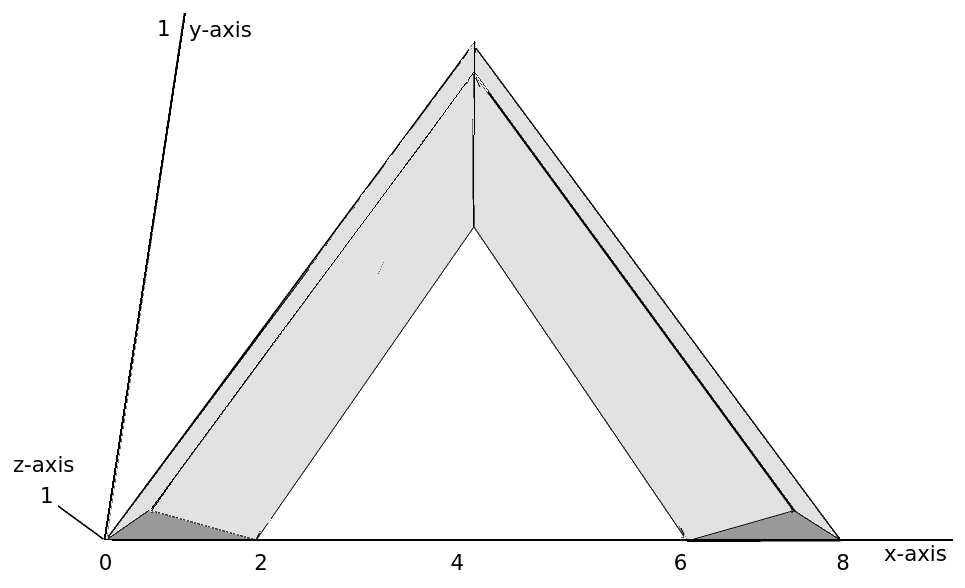}
      \label{fig:zSliceExample_GT2_set}
    }
  \subfigure[]
    {
      \includegraphics[scale=0.27]{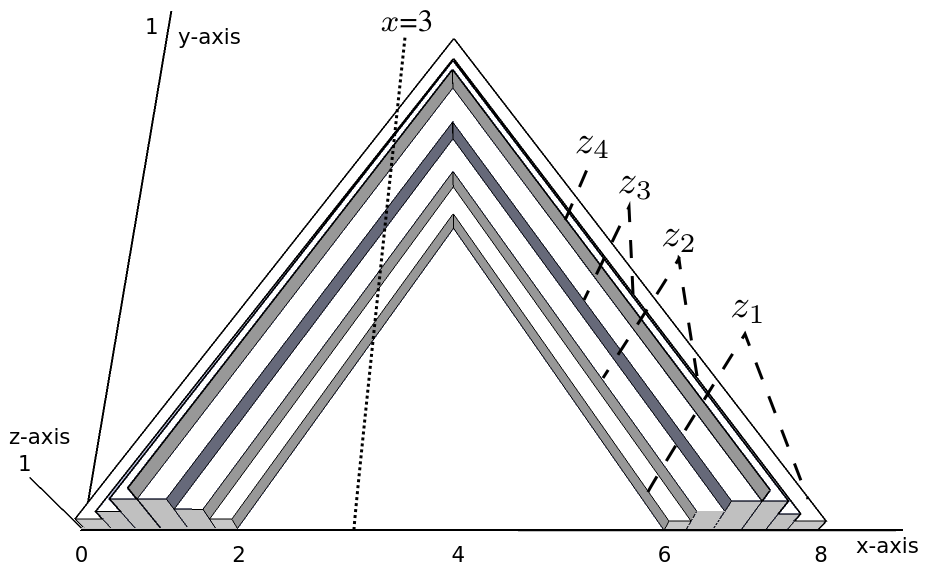}
      \label{fig:zSliceExample_zSlices_B}
    }
  \subfigure[]
    {
      \includegraphics[scale=0.26]{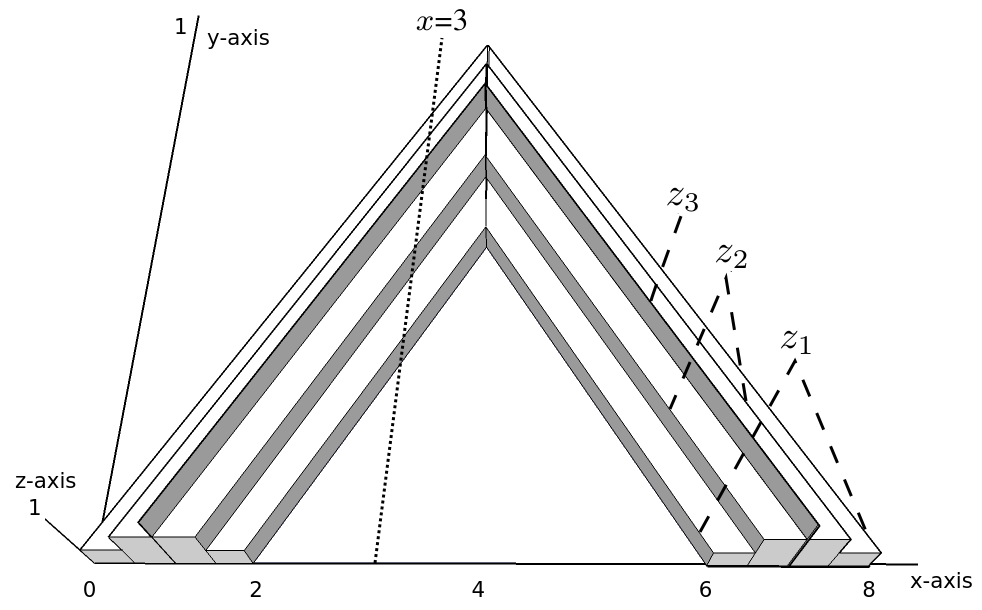}
      \label{fig:zSliceExample_zSlices_C}
    }
  \subfigure[]
    {
      \includegraphics[scale=0.20]{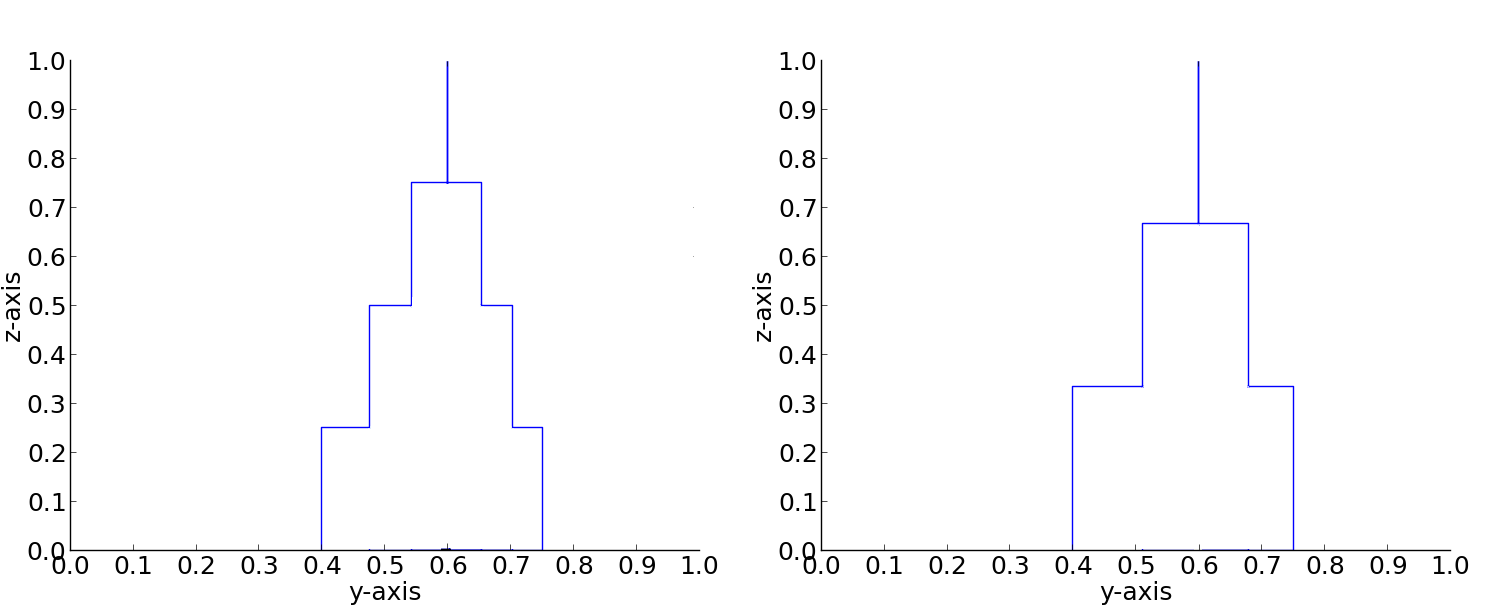}
      \label{fig:zSliceExample_vertical_slice}
    }
  \subfigure[]
    {
      \includegraphics[scale=0.20]{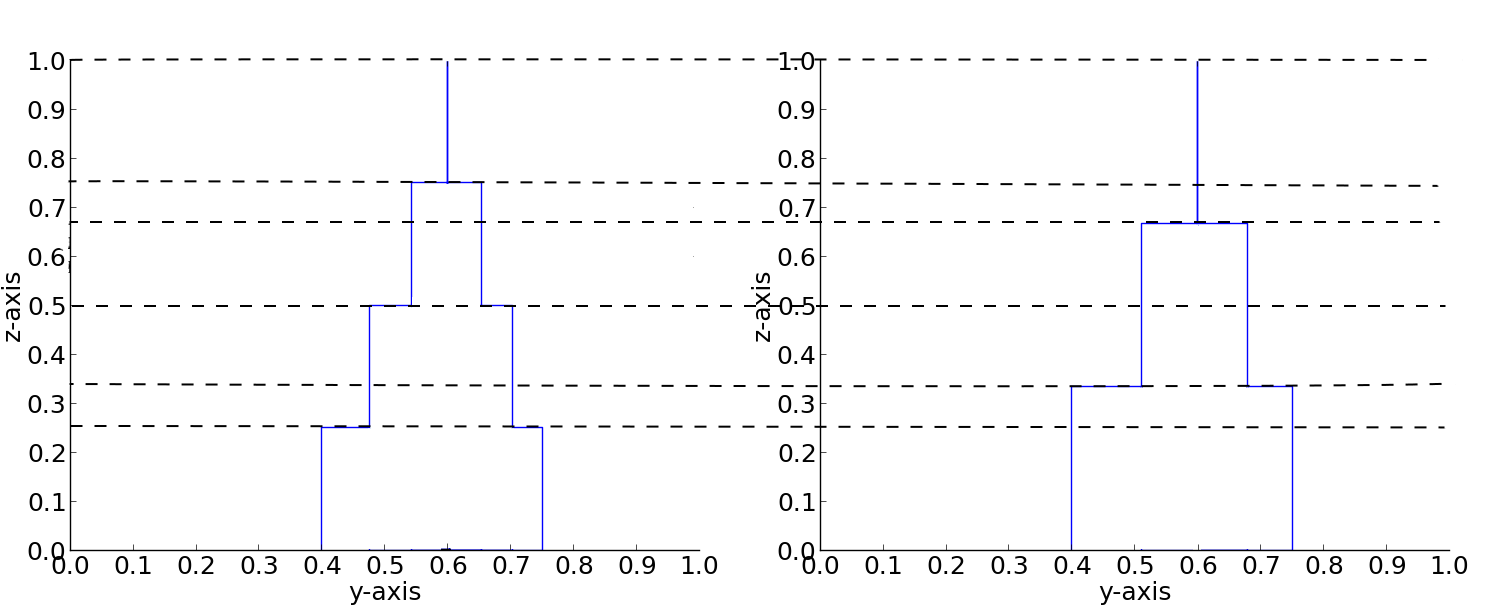}
      \label{fig:zSliceExample_vertical_slice_dashed}
    }
  \caption{(a) A general T2 FS $\tilde{A}$. (b) A zSlices-based model, $\tilde{B}$, of $\tilde{A}$ with four zLevels. (c) A zSlices-based model, $\tilde{C}$, of $\tilde{A}$ with three zLevels. (d) Vertical slices of $\tilde{B}$ and $\tilde{C}$ at $x$=3. (e) Vertical slices of $\tilde{B}$ and $\tilde{C}$ at $x$=3 with dashed lines marking their shared zLevels.}
  \label{fig:zSlices_intersection}
\end{figure}

It is worthwhile to note that in zSlices-based fuzzy logic systems each set will typically use the same number of zLevels throughout the system, however, it is possible that a SM will be required for two zSlices-based T2 FSs which use different (numbers of) zLevels. 
For example, consider the general T2 FS $\tilde{A}$ in Fig. \ref{fig:zSliceExample_GT2_set}. Two zSlices representations of this same set are shown in Fig. \ref{fig:zSliceExample_zSlices_B} and Fig. \ref{fig:zSliceExample_zSlices_C}. Set $\tilde{B}$, in Fig. \ref{fig:zSliceExample_zSlices_B}, uses four zSlices, where $z_1$ = 0.25, $z_2$ = 0.5, $z_3$ = 0.75 and $z_4$ = 1, and set $\tilde{C}$, in Fig. \ref{fig:zSliceExample_zSlices_C}, uses three zSlices, where $z_1$ = 0.33, $z_2$ = 0.66 and $z_3$ = 1. 
To clearly show the zLevels of each set, the vertical slices of $\tilde{B}$ and $\tilde{C}$ at $x=3$ are shown in Fig. \ref{fig:zSliceExample_vertical_slice}. 

The similarity of sets $\tilde{B}$ and $\tilde{C}$ is calculated using the union of their zLevels as follows: 
\begin{equation}
  L = \bigcup_{m=1}^M z_m \cup \bigcup_{n=1}^N z_n
  \label{union_of_zLevels}
\end{equation}
Where $M$ and $N$ are the number of zLevels used by each respective FS. In this example, $M=4$ and $N=3$, so $L = \{0.25, 0.5, 0.75, 1.0\} \cup \{0.33, 0.66, 1.0\} = \{0.25, 0.33, 0.5, 0.66, 0.75, 1.0\}$. 
These zLevels are shown in Fig. \ref{fig:zSliceExample_vertical_slice_dashed}, represented by dashed lines. Note that in set $\tilde{C}$ the zSlice at $z_i=0.25$ has the same FOU as the zSlice at $z_i=0.33$. 
A worked numerical example of measuring the similarity between sets with different numbers of zLevels is included in the appendix.

Through (\ref{union_of_zLevels}), (\ref{zSliceSM}) is not restricted to only zSlices-based general T2 FSs with identical or different (numbers of) zLevels, but can also be applied to zSlices-based general T2 FSs in combination with interval T2 FSs or T1 FSs. 

\subsection{Properties of the Extended Similarity Measure}
\newtheorem{thm}{\bf{Theorem}}
The following proves that when extending any interval T2 SM as shown in (\ref{zSliceSM}), all the common properties for SMs, namely, reflexivity, symmetry, transitivity, and overlapping, that hold for the original interval T2 SM will also hold in the extended zSlices-based T2 SM.

Considering the four properties of similarity introduced in Section I, an overview of the properties of each interval T2 SM introduced in Section II is presented in Table I. We consider each property below:

\begin{table}
  \centering
  \caption{An overview of the similarity properties of interval T2 SMs}
  \begin{tabular}{ c  c  c  c  c }
    \toprule
    & reflexivity & symmetry & transitivity & overlapping  \\ \midrule
    Zeng \& Li & \tick & \tick & \tick & \\ 
    Jaccard & \tick & \tick & \tick & \tick \\ 
    Gorza{\l}czany &  &  & \tick & \tick \\
    Bustince & \tick & \tick & \tick & \\ \bottomrule
  \end{tabular}
  \label{tab:propertiesTable}
\end{table}

\begin{thm}[Reflexivity]
 $S_{ZS}(\tilde{A}, \tilde{B}) = 1 \Longleftrightarrow S_{\lambda}(\tilde{A}_{z_i}, \tilde{B}_{z_i}) = 1$ for each $z \in Z$, where $Z$ is the set of all zLevels.
\end{thm}
\begin{IEEEproof}
  The similarity of each zSlice can be calculated as $S_{\lambda}(\tilde{A}_{z_i}, \tilde{B}_{z_i})$ where $\tilde{A}_{z_i}$ and $\tilde{B}_{z_i}$ are interval T2 FSs for which secondary grade is at $z_i$. If at each zLevel $\tilde{A}_{z_i}$ = $\tilde{B}_{z_i}$, then $S_{\lambda}(\tilde{A}_{z_i}, \tilde{B}_{z_i}) = 1$, and so $S_Z(\tilde{A}, \tilde{B})$ will be calculated as $\frac{\sum^I_{i=1} z_i \times 1}{\sum^I_{i=1} z_i} = 1$.
  
  Alternatively, if at any zSlice  $S_{\lambda}(\tilde{A}_{z_i}, \tilde{B}_{z_i}) < 1$, 
  then $\frac{\sum^I_{i=1} S_{\lambda}(\tilde{A}_{z_i}, \tilde{B}_{z_i})}{I} < 1$, 
  where $I$ is the total number of zSlices. Therefore
  $\frac{\sum^I_{i=1} z_i S_{\lambda}(\tilde{A}_{z_i}, \tilde{B}_{z_i})}{\sum^I_{i=1} z_i} < 1$.
  Thus, $S_{ZS}(\tilde{A}, \tilde{B}) \neq 1$ if $S_{\lambda}(\tilde{A}_{z_i}, \tilde{B}_{z_i}) \neq 1$ for any $z \in Z$.
\end{IEEEproof}

\begin{thm}[Symmetry]
 $S_{ZS}(\tilde{A}, \tilde{B}) = S_{ZS}(\tilde{B}, \tilde{A}) \Longleftrightarrow S_{\lambda}(\tilde{A}_{z_i}, \tilde{B}_{z_i}) = S_{\lambda}(\tilde{B}_{z_i}, \tilde{A}_{z_i})$ 
  for each $z \in Z$.
\end{thm}
\begin{IEEEproof}
  Observe that (\ref{zSliceSM}) does not depend on the order of $\tilde{A}$ and $\tilde{B}$, thus if $S_{\lambda}(\tilde{A}_z, \tilde{B}_z) = S_{\lambda}(\tilde{B}_z, \tilde{A}_z)$, then the same will be true for $S_Z(\tilde{A}, \tilde{B})$.
\end{IEEEproof}

\begin{thm}[Transitivity]
 $S_{ZS}(\tilde{A}, \tilde{B}) \leq S_{ZS}(\tilde{A}, \tilde{C}) \Longleftrightarrow S_{\lambda}(\tilde{A}_{z_i}, \tilde{B}_{z_i}) \leq S_{\lambda}(\tilde{A}_{z_i}, \tilde{C}_{z_i})$ 
  for each $z \in Z$.
\end{thm}
\begin{IEEEproof}
  If $S_{\lambda}(\tilde{A}_{z_i}, \tilde{B}_{z_i}) 
  \leq 
  S_{\lambda}(\tilde{A}_{z_i}, \tilde{C}_{z_i})$ 
  for each $z \in Z$ then 
  $\sum^I_{i=1} S_{\lambda}(\tilde{A}_{z_i}, \tilde{B}_{z_i}) 
  \leq 
  \sum^I_{i=1} S_{\lambda}(\tilde{A}_{z_i}, \tilde{C}_{z_i})$ 
  and therefore 
  $\frac{\sum^I_{i=1} z_i S_{\lambda}(\tilde{A}_{z_i}, \tilde{B}_{z_i})}{\sum^I_{i=1} z_i}
  \leq 
  \frac{\sum^I_{i=1} z_i S_{\lambda}(\tilde{A}_{z_i}, \tilde{C}_{z_i})}{\sum^I_{i=1} z_i}$
\end{IEEEproof}

\begin{thm}[Overlapping]
 $S_{ZS}(\tilde{A}, \tilde{B}) = 0 \Longleftrightarrow S_{\lambda}(\tilde{A}_{z_i}, \tilde{B}_{z_i}) = 0$ for each $z \in Z$.
\end{thm}
\begin{IEEEproof}
  If at each zSlice $S_{\lambda}(\tilde{A}_{z_i}, \tilde{B}_{z_i}) = 0$ then $S_Z(\tilde{A}, \tilde{B})$ will be calculated as $\frac{\sum^I_{i=1} z_i \times 0}{\sum^I_{i=1} z_i} = 0$.
  
  Alternatively, if at any zSlice  $S_{\lambda}(\tilde{A}_{z_i}, \tilde{B}_{z_i}) > 0$, 
  then $\frac{\sum^I_{i=1} S_{\lambda}(\tilde{A}_{z_i}, \tilde{B}_{z_i})}{I} > 0$, 
  where $I$ is the total number of zSlices. Therefore
  $\frac{\sum^I_{i=1} z_i S_{\lambda}(\tilde{A}_{z_i}, \tilde{B}_{z_i})}{\sum^I_{i=1} z_i} > 0$.
  Thus, $S_{ZS}(\tilde{A}, \tilde{B}) \neq 0$ if $S_{\lambda}(\tilde{A}_{z_i}, \tilde{B}_{z_i}) \neq 0$ for any $z \in Z$.
\end{IEEEproof}

\section{Demonstrations}
In this section, three demonstrations are given to present the general T2 SM. The first demonstration uses the interval T2 SMs introduced in Section II on interval T2 FSs, and demonstrations 2 and 3 apply the general T2 SM on different general T2 FSs. Reviews of the interval T2 measures introduced have been presented by Wu and Mendel in \cite{WuVector} and \cite{WuComparative}, however, a brief review is given in this section so that comparisons can be made against the demonstrations for the zSlices-based general T2 case. 

\subsection{Demonstration 1 - Comparison of Interval Type-2 Approaches}
\begin{figure}
  \centering
    \includegraphics[scale=0.47]{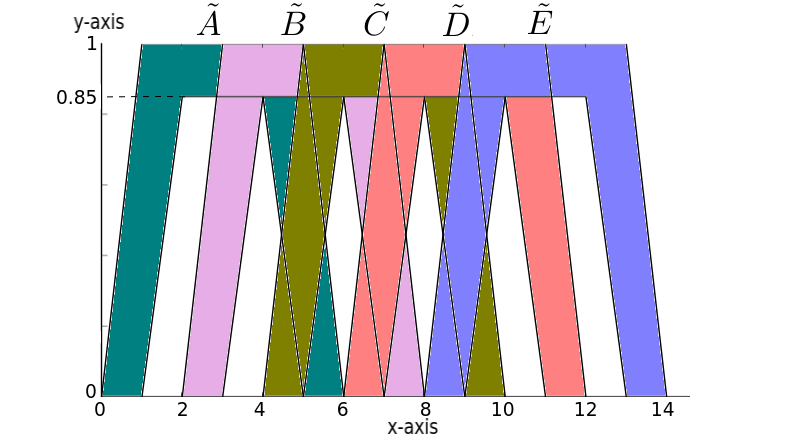}
    \caption{Trapezoidal interval T2 FSs used to test interval T2 SMs.}
    \label{fig:trapezoid_fuzzy_sets}
\end{figure}

In this demonstration, each method was applied to the interval T2 FSs displayed in Fig. \ref{fig:trapezoid_fuzzy_sets}, the results of which are shown in Table \ref{tab:trapezoid_results}. The $x$-axis was discretised into 100 equally distanced points, and minimum t-norm was used for Bustince's SM. 
As in \cite{WuVector}, it can be observed that neither Zeng and Li's nor Bustince's measures support the property of overlapping. When the FSs being measured are disjoint, Zeng and Li's SM increases as the distance between the sets increases and Bustince's measure always gives a constant non-zero value. This can be seen in the results of sets $\tilde{A}$ and $\tilde{D}$ and of sets $\tilde{A}$ and $\tilde{E}$. Depending on the application, this may not be what is expected as it is often presumed that $S(\tilde{A}, \tilde{B})$ either decreases as the distance between $\tilde{A}$ and $\tilde{B}$ increases, or is given as 0. Additionally, Gorza{\l}czany's measure has given (1.0, 1.0) for sets $\tilde{A}$ and $\tilde{B}$ because this measure will always give (1.0 1.0) when $max_{x \in X}\underline{\mu}_{\tilde{A}}(x) = max_{x \in X}\underline{\mu}_{\tilde{B}}(x)$ and $max_{x \in X}\overline{\mu}_{\tilde{A}}(x) = max_{x \in X}\overline{\mu}_{\tilde{B}}(x)$; as is true for sets $\tilde{A}$ and $\tilde{B}$. Jaccard's SM, however, gives expected results. 

\begin{table}[h!]
\caption{Comparison of interval T2 SMs using the FSs displayed in Fig. \ref{fig:trapezoid_fuzzy_sets}}
  \begin{center}
    \begin{tabular}{  c  c  c  c  c  c  }
      \toprule
      & $S(\tilde{A}, \tilde{A})$ & $S(\tilde{A}, \tilde{B})$ & $S(\tilde{A}, \tilde{C})$ & $S(\tilde{A}, \tilde{D})$ & $S(\tilde{A}, \tilde{E})$ \\ \midrule
      Zeng \& Li & 1.0 & 0.538 & 0.345 & 0.371 & 0.461 \\ 
      Jaccard    & 1.0 & 0.342 & 0.071 & 0.0 & 0.0 \\ 
      Gorza{\l}czany & (1.0, 1.0) & (1.0, 1.0) & (0.0, 1.0) & (0.0, 0.0) & (0.0, 0.0) \\ 
      Bustince & (1.0, 1.0) & (0.0, 0.15) & (0.0, 0.15) & (0.0, 0.15) & (0.0, 0.15) \\ \bottomrule
    \end{tabular}
  \end{center}
  \label{tab:trapezoid_results}
\end{table}

\subsection{Demonstrations of zSlices-Based General Type-2 Approaches}
Two different demonstrations (demonstrations 2 and 3) apply the zSlices-based SM to zSlices-based general T2 FSs. Examples of trapezoidal and triangular FSs are presented, however the SM will work for any type of sets. These demonstrations use the SMs by Zeng and Li, Jaccard, Gorza{\l}czany and Bustince; as defined in (\ref{zeng}), (\ref{jaccard}), (\ref{Gorz}) and (\ref{Bustince}), respectively. Demonstration 2 employs the zSlices-based SM on trapezoidal general T2 FSs that are based on the interval T2 FSs used in demonstration 1. This enables comparisons to be made between the interval T2 SMs and their extensions. The final demonstration measures the similarity between triangular T2 FSs, providing an additional example to demonstrate the zSlices-based SMs.

\subsubsection{Demonstration 2}
\begin{figure}
  \centering
    \subfigure[]
    {
      \includegraphics[scale=0.34]{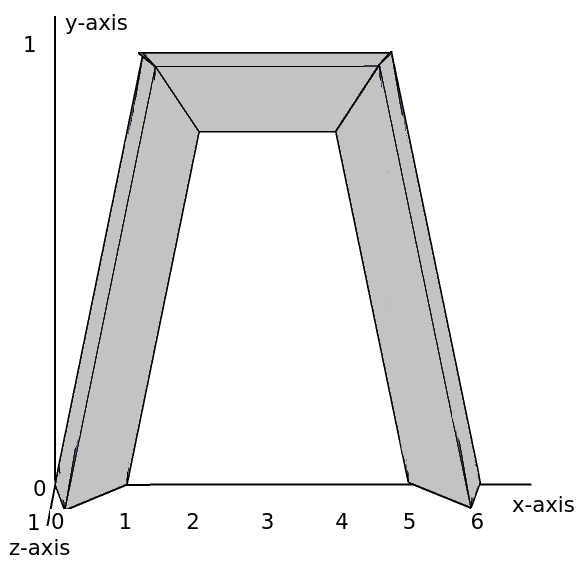}
      \label{fig:zSlice_trapezoid_GT2FS}
    }
    \subfigure[]
    {
      \includegraphics[scale=0.38]{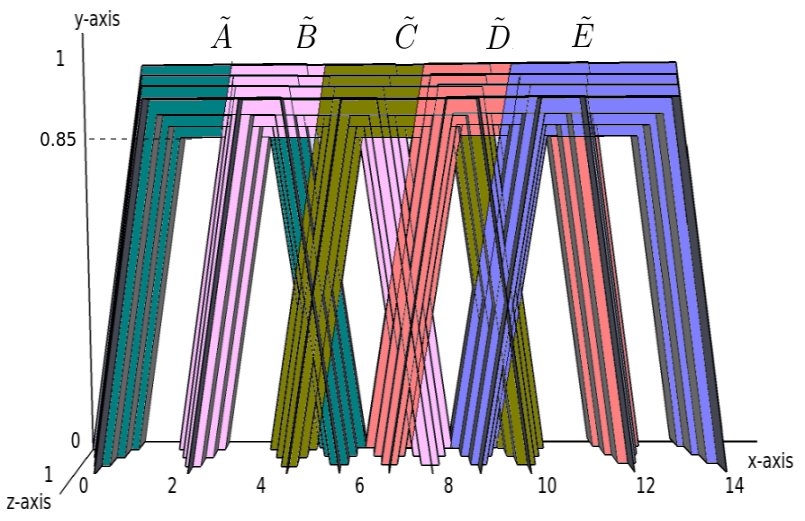}
      \label{fig:zSlice_trapezoid_ZSlices_2D}
    }
    \subfigure[]
    {
      \includegraphics[scale=0.34]{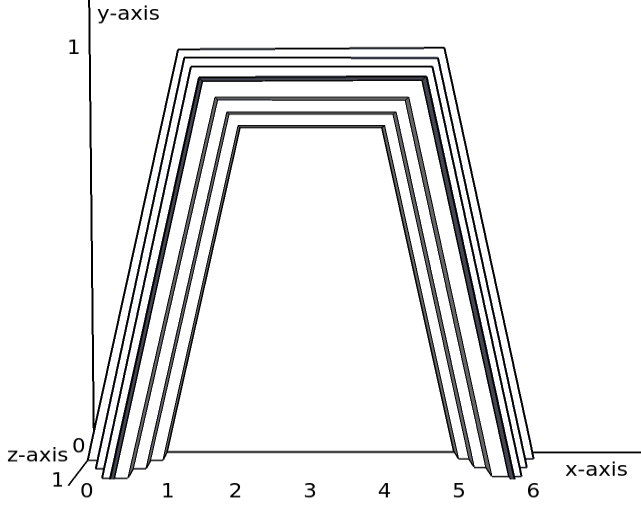}
      \label{fig:zSlice_trapezoid_single_set_front}
    }
    \subfigure[]
    {
      \includegraphics[scale=0.34]{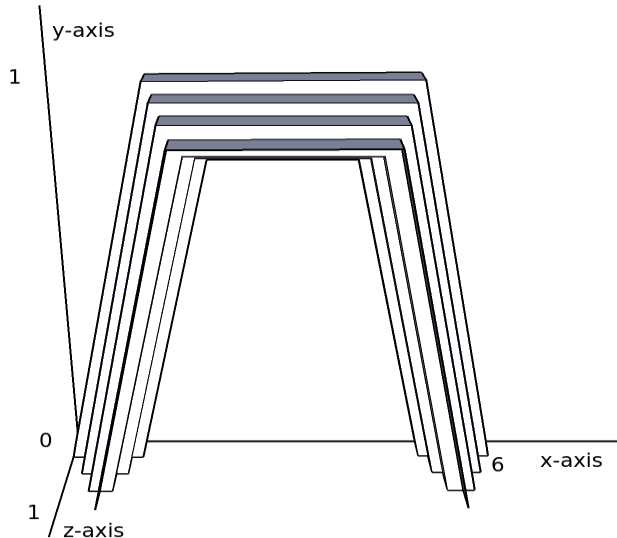}
      \label{fig:zSlice_trapezoid_single_set_above}
    }
  \caption{(a) A trapezoidal general T2 FS based on the FSs in Fig \ref{fig:trapezoid_fuzzy_sets}. (b) Five zSlices-based representations of (a) using four zSlices arranged as Fig \ref{fig:trapezoid_fuzzy_sets}. (c) Front view of set $\tilde{A}$. (d) Tilted view of set $\tilde{A}$.}
\end{figure}

For this demonstration, each interval T2 FS in Fig. \ref{fig:trapezoid_fuzzy_sets} was altered into a standard general T2 FS with a principal MF as shown in Fig. \ref{fig:zSlice_trapezoid_GT2FS}, which displays the set $\tilde{A}$. Each set has a triangular secondary MF, the peak of which is at the centre of the FOU. The FSs were next represented by zSlices-based T2 FSs with four zLevels at coordinates 0.25, 0.5, 0.75 and 1.0. Note that the number of zLevels was chosen as four to provide a clear demonstration, however more zLevels may be used to gain a more accurate representation of the original general T2 FS. The zSlices of all sets are shown in Fig. \ref{fig:zSlice_trapezoid_ZSlices_2D}. Fig. \ref{fig:zSlice_trapezoid_single_set_front} and Fig. \ref{fig:zSlice_trapezoid_single_set_above} show the zSlices representation of the FS $\tilde{A}$ in more detail. The results of this experiment, for which the $x$-axis was discretised into 100 equally distanced points, are shown in Table \ref{tab:trap_zslices_results}.

Comparing the results from the experiments on the interval T2 FSs in Table \ref{tab:trapezoid_results} with the zSlices-based general T2 FSs in Table \ref{tab:trap_zslices_results}, each method has maintained the same similarity properties shown in Table \ref{tab:propertiesTable} in both experiments as proven in Section III B. However, though one would assume that the results of each experiment should be the same, the measurements given by each method have changed slightly. We next address the altered result of each SM.

\begin{table}[h!]
\caption{Comparison of the zSlices similarity measure using interval T2 SMs on the zSlices-based T2 FSs displayed in Fig. \ref{fig:zSlice_trapezoid_ZSlices_2D}.}
  \begin{center}
    \begin{tabular}{  c  c  c  c  c  c  }
      \toprule
      & $S(\tilde{A}, \tilde{A})$ & $S(\tilde{A}, \tilde{B})$ & $S(\tilde{A}, \tilde{C})$ & $S(\tilde{A}, \tilde{D})$ & $S(\tilde{A}, \tilde{E})$ \\ \midrule
      Zeng \& Li & 1.0 & 0.496 & 0.267 & 0.345 & 0.443 \\ 
      Jaccard & 1.0 & 0.335 & 0.041 & 0.0 & 0.0 \\ 
      Gorza{\l}czany & (1.0, 1.0) & (1.0, 1.0) & (0.33, 0.66) & (0.0, 0.0) & (0.0, 0.0) \\ 
      Bustince & (1.0, 1.0) & (0.05, 0.1) & (0.05, 0.1)  & (0.05, 0.1) & (0.05, 0.1) \\ \bottomrule
    \end{tabular}
  \end{center}
  \label{tab:trap_zslices_results}
\end{table}

\begin{figure}
  \centering
    \subfigure[]
    {
      \includegraphics[scale=0.35]{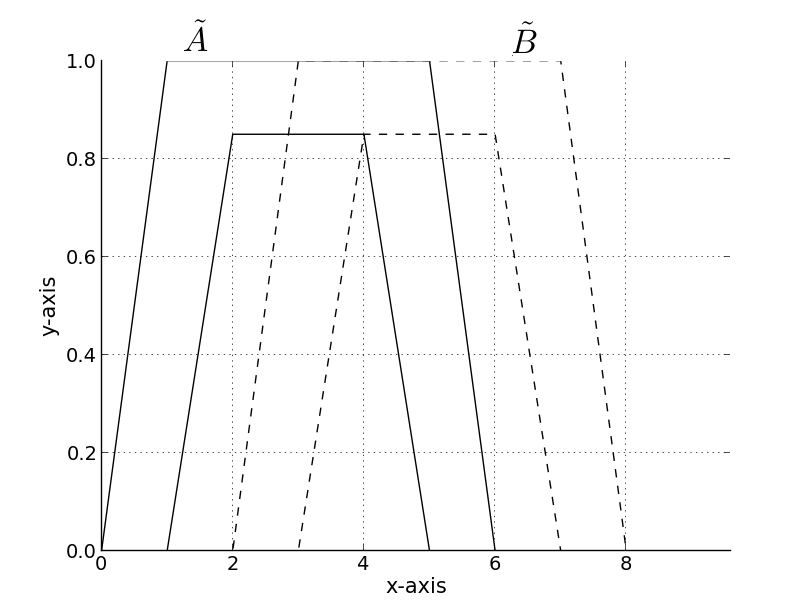}
    }
    \hfil
    \subfigure[]
    {
      \includegraphics[scale=0.35]{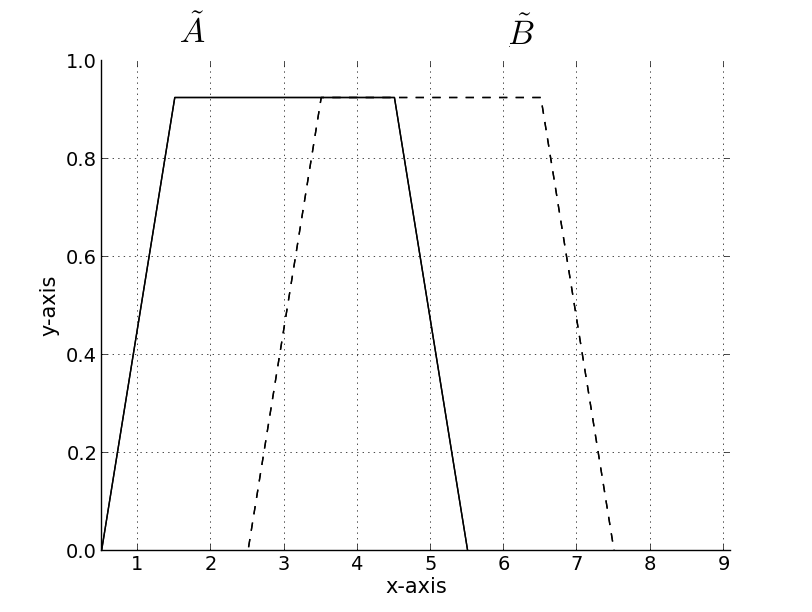}
    }
  \caption{Sets $\tilde{A}$ and $\tilde{B}$ from Fig. \ref{fig:zSlice_trapezoid_ZSlices_2D} at different zLevels. (a) $z_i$ = 0.25. (b) $z_i$ = 1. Note the intersections of the upper and lower MF of $\tilde{A}$ and $\tilde{B}$ have changed at each zLevel.}
  \label{fig:discussion_sets}
\end{figure}

\begin{itemize}
\item{Zeng and Li} \\
As in Table \ref{tab:trapezoid_results}, Zeng and Li's measure follows the same trait in which the SM increases as the distance between two disjoint sets increases. However, the measurements between the interval T2 FSs and the zSlices-based general T2 FSs have changed slightly. This is because Zeng and Li's measure focuses on the distance between the upper and lower MFs of each set, the positions of which alters at each zLevel. For example, Fig. \ref{fig:discussion_sets} shows sets $\tilde{A}$ and $\tilde{B}$ at zLevels 0.25 and 1 ($z_i = 0.25$ represents the sets each as an interval T2 FS in which the upper and lower MFs are unchanged from Fig. \ref{fig:trapezoid_fuzzy_sets}, and $z_i = 1$ represents the 
sets each as a T1 FS). It can be seen that the distance between the upper and lower MFs of each set alters when the zLevel alters, which affects the result of the SM.

\item{Jaccard} \\
The results of Jaccard's SM also differ Table \ref{tab:trapezoid_results}. This is for the same reason that Zeng and Li's measurements changed, which is that Jaccard's measure focuses on the distance between the upper and lower MFs of each set, which alters at each zLevel.

\item{Gorza{\l}czany} \\
As in Table \ref{tab:trapezoid_results}, Gorza{\l}czany's measure still has the property of overlapping and has also given (1.0, 1.0) for the trapezoidal zSlice-based general T2 FSs $\tilde{A}$ and $\tilde{B}$ because $max_{x \in X}\underline{\mu}_{\tilde{A}}(x) = max_{x \in X}\underline{\mu}_{\tilde{B}}(x)$ and $max_{x \in X}\overline{\mu}_{\tilde{A}}(x) = max_{x \in X}\overline{\mu}_{\tilde{B}}(x)$ is true for each zSlice. Gorza{\l}czany's results for the zSlices experiment differ from the interval T2 case because the method focuses on the coordinates at which the two sets intersect, which changes at each zLevel. Referring to Fig. \ref{fig:discussion_sets}, the intersection of the lower and upper MFs of $\tilde{A}$ and $\tilde{B}$ when $z_i$=0.25 are at $y$=0.85 and $y$=1, respectively. However, the intersections at $z_i$=1 are at y=0.925 for both the lower and upper MFs. It is because these coordinates change that the results of the SM have also changed.

\item{Bustince} \\
As in the demonstration for interval T2 FSs shown in Table \ref{tab:trapezoid_results}, this SM gives a non-zero value when the sets being measured are disjoint. Finally, Bustince's results have also changed compared to the results in Table \ref{tab:trapezoid_results} because this method focuses on the specific position of each FS's upper and lower MF; which, as stated earlier, changes at each zLevel. 
\end{itemize}

\subsubsection{Demonstration 3}
\begin{figure}
  \centering
    \subfigure[]
    {
      \includegraphics[scale=0.32]{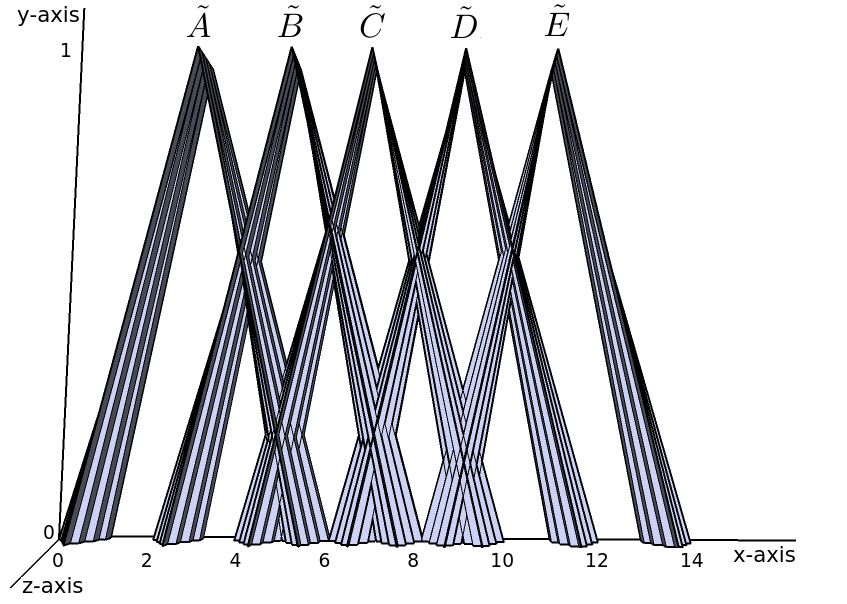}
    }
    \subfigure[]
    {
      \includegraphics[scale=0.34]{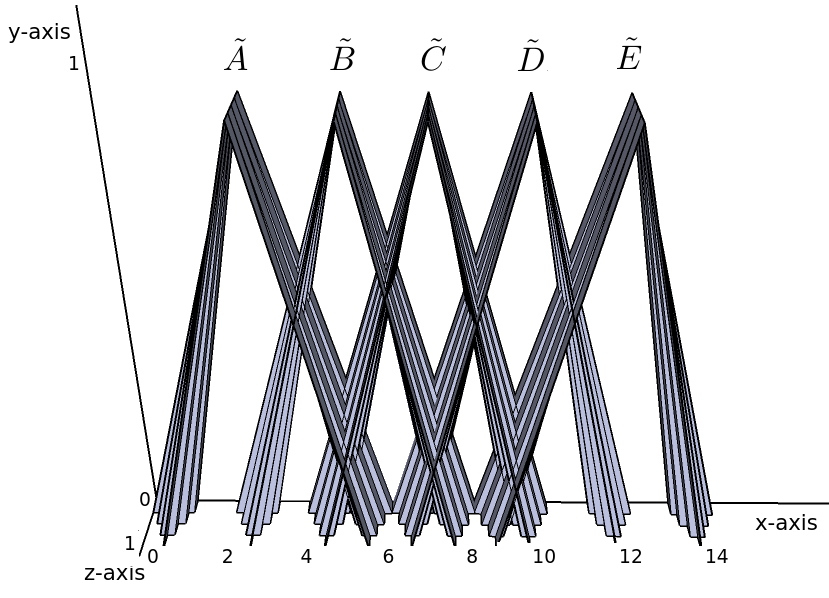}
    }
  \caption{zSlices-based triangular fuzzy sets with four zLevels used in experiment 2. (a) Front view of fuzzy sets. (b) Tilted view fuzzy sets.}
  \label{fig:tri_zSLices_fuzzy_sets}
\end{figure}

The FSs used in this demonstration are shown in Fig. \ref{fig:tri_zSLices_fuzzy_sets}. They are based on general T2 FSs in which the secondary MF is triangular with the peak at the centre of the FOU. The results of this experiment are shown in Table \ref{tab:tri_zslices_results}. For each method the $x$-axis was discretised into 100 equally distanced points, and the $z$-axis was discretised into four zLevels at coordinates 0.25, 0.5, 0.75 and 1.0. The results shown in Table \ref{tab:tri_zslices_results} correspond with the results in Table \ref{tab:trapezoid_results} and Table \ref{tab:trap_zslices_results} and each method has retained the similarity properties shown in Table \ref{tab:propertiesTable} and proven is Section III B.

In this demonstration, the values from Jaccard's SM have declined more rapidly than in the previous demonstration, whilst Zeng \& Li's, and Bustince's SMs give similar values to Table \ref{tab:trap_zslices_results}. Gorza{\l}czany's SM has gradually declined as the distance between the peak of each set increases. In this case the SM has also shown reflexivity due to the shape and position of the sets.

\begin{table}[h!]
\caption{Comparison of zSlices SM using interval T2 SMs on zSlices-based general T2 FSs displayed in Fig. \ref{fig:tri_zSLices_fuzzy_sets}.}
  \begin{center}
    \begin{tabular}{  c  c  c  c  c  c  }
      \toprule
      & $S(\tilde{A}, \tilde{A})$ & $S(\tilde{A}, \tilde{B})$ & $S(\tilde{A}, \tilde{C})$ & $S(\tilde{A}, \tilde{D})$ & $S(\tilde{A}, \tilde{E})$ \\ \midrule
      Zeng \& Li & 1.0 & 0.565 & 0.49 & 0.56 & 0.626 \\
      Jaccard & 1.0 & 0.221 & 0.024 & 0.0 & 0.0 \\
      Gorza{\l}czany & (1.0, 1.0) & (0.58, 0.63) & (0.14, 0.25) & (0.0, 0.0) & (0.0, 0.0) \\ 
      Bustince & (1.0, 1.0) & (0.14, 0.25) & (0.01, 0.01)  & (0.01, 0.01) & (0.01, 0.01) \\ \bottomrule
    \end{tabular}
  \end{center}
  \label{tab:tri_zslices_results}
\end{table}

\section{Conclusion}
In this paper we have introduced a general method of extending existing SMs on interval T2 FSs to SMs on general T2 FSs through the use of the zSlices based general T2 FS representation \cite{zSlicesPaper2} . We have shown how, based on this approach, any interval T2 FS based SM can be extended to the general T2 case and that the extension preserves all the common initial properties for SMs of the interval T2 case, namely reflexivity, symmetry, transitivity and overlapping.

We have demonstrated the method by extending a series of the most common SMs for interval T2 FSs to the general T2 case and providing examples and comparisons of applying them based on different types (i.e. triangular and trapezoidal) of FSs.

SMs provide an essential tool for the reasoning on FSs. In the future we plan to deploy the developed similarity measures for general T2 FSs in a variety of applications with a specific focus on Computing With Words \cite{CWW}, where interval and general T2 FSs provide a promising avenue for capturing subjective concept and word models \cite{mendel2010perceptual, Miller2012} and SMs provide an essential tool for reasoning.

\bibliographystyle{IEEEtran}
\bibliography{IEEEabrv.bib,papers.bib}
\enlargethispage{-0.28in}
\appendix[Numerical Example for the zSlices-based General Type-2 Similarity Measure]
The proposed zSlices-based SM is demonstrated using Jaccard's interval T2 SM, as shown in (\ref{jaccard}) \cite{WuComparative}. Jaccard's measure has been chosen because it possesses all four properties of similarity shown earlier and requires only a single step of calculation for each zSlice, thus it is a favourable method for demonstrating the extension.

Consider the general T2 FS $\tilde{A}$ in Fig. \ref{fig:zSliceExample_GT2_set}. Two zSlices representations of this set are shown in Fig. \ref{fig:zSliceExample_zSlices_B} and Fig. \ref{fig:zSliceExample_zSlices_C}. Set $\tilde{B}$, in Fig. \ref{fig:zSliceExample_zSlices_B}, consists of four zSlices, the zLevels of which are at $z_1$ = 0.25, $z_2$ = 0.5, $z_3$ = 0.75 and $z_4$ = 1, and set $\tilde{C}$, in Fig. \ref{fig:zSliceExample_zSlices_C}, consists of three zSlices, the zLevels of which are at  $z_1$ = 0.33, $z_2$ = 0.66, $z_3$ = 1. To clearly show the zLevels of each set, the vertical slice of sets $\tilde{B}$ and $\tilde{C}$ at $x=3$ is shown in Fig. \ref{fig:zSliceExample_vertical_slice}. 

Due to each set using different numbers of zLevels, the combination of their zLevels will need to be used to calculate their similarity as shown in (\ref{union_of_zLevels}). Therefore, the zLevels used will be $z_1$ = 0.25, $z_2$ = 0.33, $z_3$ = 0.5, $z_4$ = 0.66, $z_5$ = 0.75 and $z_6$ = 1.0. 
The similarity of the FSs $\tilde{B}$ and $\tilde{C}$ is calculated using the zSlices SM in (\ref{zSliceSM}) with Jaccard's interval T2 SM in (\ref{jaccard}) as follows:
\begin{equation}
  \begin{array}{l}
  S_{ZS}(\tilde{B},\tilde{C}) = \\
    \frac
      {\sum_{i \in L}  z_i  
	\frac{
	  \int_X min(\overline{\mu}_{\tilde{B}_{z_i}}(x), \overline{\mu}_{\tilde{C}_{z_i}}(x)) dx + 
	  \int_X min(\underline{\mu}_{\tilde{B}_{z_i}}(x), \underline{\mu}_{\tilde{C}_{z_i}}(x)) dx}
	  {
	  \int_X max(\overline{\mu}_{\tilde{B}_{z_i}}(x), \overline{\mu}_{\tilde{C}_{z_i}}(x)) dx + 
	  \int_X max(\underline{\mu}_{\tilde{B}_{z_i}}(x), \underline{\mu}_{\tilde{C}_{z_i}}(x)) dx}}
      {\sum_{i \in L}  z_i}
  \end{array}
\end{equation}
where $L=\{0.25, 0.33, 0.5, 0.66, 0.75, 1\}$.

The FSs $\tilde{B}$ and $\tilde{C}$ are discretised as follows; for space consideration we focus on the left side of the symmetrical sets:
\begin{equation}
 \nonumber
\begin{array}{l @{\hspace{2bp}} l}
\mu_{\tilde{B}}(x=1) = & 1/0 + 0.75/0.093 + 0.5/0.176 + 0.25/0.25 \\
\mu_{\tilde{B}}(x=2) = & 0.25/0 + 0.5/0.119 + 0.75/0.217 + 1/0.3 \\
		       & +\ 0.75/0.373 + 0.5/0.439 + 0.25/0.5 \\
\mu_{\tilde{B}}(x=3) = & 0.25/0.4 + 0.5/0.476 + 0.75/0.542 + 1/0.6 \\
		       & +\ 0.75/0.653 + 0.5/0.703 + 0.25/0.75 \\
\mu_{\tilde{B}}(x=4) = & 0.25/0.8 + 0.5/0.833 + 0.75/0.867 + 1/0.9 \\
			& +\ 0.75/0.933 + 0.5/0.967 + 0.25/1 \\ \\
\mu_{\tilde{C}}(x=1) = & 1/0 + 0.66/0.136 + 0.33/0.25 \\
\mu_{\tilde{C}}(x=2) = & 0.33/0 + 0.66/0.17 + 1/0.3 \\ & +\ 0.66/0.407 + 0.33/0.5 \\
\mu_{\tilde{C}}(x=3) = & 0.33/0.4 + 0.66/0.51 + 1/0.6 \\ & +\ 0.66/0.679 + 0.33/0.75 \\
\mu_{\tilde{C}}(x=4) = & 0.33/0.8 + 0.66/0.85 + 1/0.9 \\ & +\ 0.66/0.95 + 0.33/1 \\
\end{array}
\end{equation}

At $z_i$ = 0.25, Jaccard's SM is calculated as
\small
\begin{equation}
 \nonumber
 \begin{array}{l @{\hspace{2bp}} l}
 S_J(\tilde{B}_{z_1}, \tilde{C}_{z_1}) &
	  = \frac{ (0.25+0.5+0.75+1) + (0+0+0.4+0.8) }{ (0.25+0.5+0.75+1) + (0+0+0.4+0.8) } \\
      & = \frac{3.7}{3.7} = 1
  \end{array}
\end{equation}
\normalsize

At $z_i$ = 0.33,
\small
\begin{equation}
 \nonumber
 \begin{array}{l @{\hspace{2bp}} l}
 S_J(\tilde{B}_{z_2}, \tilde{C}_{z_2}) &
	  = \frac{ (0.176+0.439+0.703+0.967) + (0+0+0.4+0.8) }
		 { (0.25+0.5+0.75+1) + (0+0.119+0.476+0.833) } \\
      & = \frac{3.485}{3.928} = 0.887
 \end{array}
\end{equation}
\normalsize

At $z_i$ = 0.5,
\small
\begin{equation}
 \nonumber
 \begin{array}{l @{\hspace{2bp}} l}
 S_J(\tilde{B}_{z_3}, \tilde{C}_{z_3}) &
	  = \frac{ (0.136+0.407+0.679+0.95) + (0+0.119+0.476+0.833) }
		 { (0.176+0.439+0.703+0.967) + (0+0.17+0.51+0.85) } \\
      & = \frac{3.6}{3.815} = 0.944
  \end{array}
\end{equation}
\normalsize

At $z_i$ = 0.66,
\small
\begin{equation}
 \nonumber
 \begin{array}{l @{\hspace{2bp}} l}
 S_J(\tilde{B}_{z_4}, \tilde{C}_{z_4}) &
	  = \frac{ (0.093+0.373+0.653+0.933)+(0+0.17+0.51+0.85) }
		 { (0.136+0.407+0.679+0.95)+(0+0.217+0.542+0.867) } \\
      & = \frac{3.582}{3.798} = 0.943
  \end{array}
\end{equation}
\normalsize

At $z_i$ = 0.75,
\small
\begin{equation}
 \nonumber
 \begin{array}{l @{\hspace{2bp}} l}
 S_J(\tilde{B}_{z_5}, \tilde{C}_{z_5}) &
	= \frac{ (0+0.3+0.6+0.9) + (0+0.217+0.542+0.867) }
	      { (0.093+0.373+0.653+0.933) + (0+0.3+0.6+0.9) } \\
      & = \frac{3.426}{3.852} = 0.889
  \end{array}
\end{equation}
\normalsize

At $z_i$ = 1,
\small
\begin{equation}
 \nonumber
 \begin{array}{l @{\hspace{2bp}} l}
 S_J(\tilde{B}_{z_6}, \tilde{C}_{z_6}) &
	= \frac{(0+0.3+0.6+0.9)+(0+0.3+0.6+0.9)}{(0+0.3+0.6+0.9)+(0+0.3+0.6+0.9)} \\
      & = \frac{3.6}{3.6} = 1
  \end{array}
\end{equation}
\normalsize

Finally, combining these results in the zSlices SM gives
\small
\begin{equation}
  \nonumber
  \begin{array}{l @{\hspace{2bp}} l}
  Z_{ZS} & = \frac{0.25 \times 1 + 0.33 \times 0.887 + 0.5 \times 0.944  
		    + 0.66 \times 0.943 + 0.75 \times 0.889 + 1.0 \times 1}
	      {0.25 + 0.33 + 0.5 + 0.66 + 0.75 + 1.0} \\
  & = \frac{3.304}{3.49} = 0.947
  \end{array}
\end{equation}
\normalsize
In this example the results at $z_i$ = 0.25 and $z_i$ = 1 are both 1 because sets $\tilde{B}$ and $\tilde{C}$ are based on the same original general T2 FS and these specific zSlices are in fact equal, i.e., the positions of the lower and upper MFs are the same at the lowest zLevel - which is the same as the original set, and at the highest zLevel - which is the T1 representation of the original set. However, results from zLevels $z_i$ = 0.33 to $z_i$ = 0.75 are less than 1 because $\tilde{B}$ and $\tilde{C}$ are discretised differently along the $z$-axis and are thus not identical.
\newpage
\end{document}